\documentclass[letterpaper]{article} 
\usepackage[]{aaai2026}
\nocopyright
\usepackage{times}  
\usepackage{helvet}  
\usepackage{courier}  
\usepackage[hyphens]{url}  
\usepackage{graphicx} 
\usepackage{natbib}  
\usepackage{caption} 
\usepackage{algorithm}
\usepackage{algorithmic}

\usepackage{amsmath}
\usepackage{cleveref}

\usepackage{newfloat}
\usepackage{listings}
\DeclareCaptionStyle{ruled}{labelfont=normalfont,labelsep=colon,strut=off} 
\floatstyle{ruled}
\newfloat{listing}{tb}{lst}{}
\floatname{listing}{Listing}
\title{Semantic Similarity is a Spurious Measure of Comic Understanding: \\Lessons Learned from Hallucinations in a Benchmarking Experiment}

\author {
    Christopher Driggers-Ellis\textsuperscript{\rm 1},
    Nachiketh Tibrewal\textsuperscript{\rm 1},
    Rohit Bogulla\textsuperscript{\rm 1},
    Harsh Khanna\textsuperscript{\rm 1},\\
    Sangpil Youm\textsuperscript{\rm 1},
    Christan Grant\textsuperscript{\rm 1},
    Bonnie Dorr\textsuperscript{\rm 1}
}
\affiliations {
    \textsuperscript{\rm 1}University of Florida, Gainesville, FL 32611 \\
    \{driggersellis.cw, n.tibrewal, rbogulla, khanna.harsh, youms, christan, bonniejdorr \}@ufl.edu
}

\usepackage{bibentry}

\begin{document}

\maketitle

\begin{abstract}
A system that enables blind or visually impaired users to access comics/manga would introduce a new medium of storytelling to this community.
However, no such system currently exists.
Generative vision-language models (VLMs) have shown promise in describing images and understanding comics, but most research on comic understanding is limited to panel-level analysis.
To fully support blind and visually impaired users, greater attention must be paid to page-level understanding and interpretation.
In this work, we present a preliminary benchmark of VLM performance on comic interpretation tasks.
We identify and categorize hallucinations that emerge during this process, organizing them into generalized object-hallucination taxonomies.
We conclude with guidance on future research, emphasizing hallucination mitigation and improved data curation for comic interpretation.
\end{abstract}


\section{Introduction}

Comic books have been a fixture of American popular culture for 
over a century~\cite{rhoades2008complete}. US sales of digital comics, particularly manga, are expected to grow rapidly in coming years~\cite{gvr2024comic, gvr2024manga}. While general audiences are 
eager and able to consume more comic/manga content, these media remain 
inaccessible to blind and visually impaired users.
This audience requires a specialized screen reader 
to interpret and describe visual content.

\begin{figure}
\begingroup 
\setlength{\tabcolsep}{0pt} 
\hspace*{-.05in}
\begin{tabular}{ll}
  \includegraphics[width=0.51\columnwidth
  ]{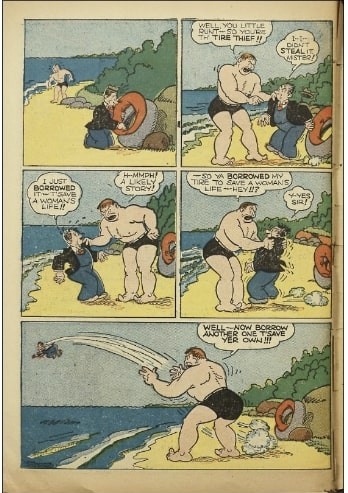}&
    \includegraphics[width=0.51\columnwidth
  ]{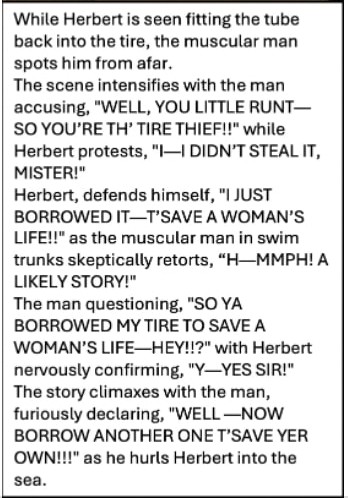}
\end{tabular}
\endgroup
  \caption{
  An image from our benchmarking corpus (left) paired with its ground truth interpretation (right).}
  \label{fig:dataset-example}
\end{figure}

While existing screen readers effectively handle web content, they fall short when it comes to complex visual storytelling. To bridge this gap, there is a need for a tool that can convey curated narrative descriptions---like
the ground truth shown in~\Cref{fig:dataset-example}---through audio. 
Yet, to our knowledge, no screen reader offers functionality tailored to
the unique demands of comics and manga. 

At the same time, vision language models (VLMs), which analyze images and 
generate
text, 
are advancing rapidly in 
reasoning ability~\cite{chow2025physbench},
with frequent improvements reported across 
computer vision (CV) and reasoning tasks.
These two abilities are exactly 
what is needed
to interpret comic books and manga.
Computer vision tasks are crucial for extracting visual features from comic pages, and tying features together into a coherent textual narrative is a reasoning task similar to image captioning.

Prior comic understanding investigations have assessed VLMs' interpretation capabilities at the panel level 
(individual images on a comic book page)~\cite{vivoli2024comicapvlmspipelinedense} and 
in short-form comics such as newspaper comic 
strips~\cite{Ramaprasad2023ComicsFE}. However, the literature lacks 
holistic studies of page-level understanding or interpretation.
The gap
is concerning, 
as page-level interpretation is key to
generating coherent text that enables screen readers to narrate
long-form comics to visually impaired users.

Most prior work 
approaches comic understanding from
a CV perspective 
rather than a natural language processing (NLP) perspective,
often using
deep learning for panel segmentation and convolutional neural networks for character and text detection in comic/manga images \citep{nguyen2018digital, augereau2018survey, multimedia_aizawa_2020}.
Few studies have linked CV methods with
comic understanding 
in the text modality~\cite{vivoli2024comicapvlmspipelinedense, Ramaprasad2023ComicsFE}. This gap 
complicates the task, as
researchers and developers must 
establish how to generate high-quality \emph{textual} narrations---e.g., via vision-language models (VLMs)---following upstream CV analysis, to produce a usable comic interpretation feature.

Recognizing these gaps as barriers
to developing 
a comic interpretation feature 
for blind and visually impaired users,
we 
present a benchmark for VLMs on 
this task.
We 
examine hallucinations, as these errors in text and visual generation severely impact the accuracy and coherence of comic interpretations. 
Using 
taxonomies from picture captioning literature to categorize hallucinations, we identify common errors such as object, relation, and attribute hallucinations.
Our exploration reveals that
the benchmark results may be unreliable. We therefore
refocus and make the following contributions to the comic understanding literature:
\begin{itemize}
    \item Evidence that simple semantic similarity metrics are unsuitable for comic interpretation and related tasks.
    \item Investigation of common VLM hallucinations during comic interpretation.
    \item Categorization of common hallucinations using existing VLM taxonomies.
\end{itemize}

We review prior work on comic understanding and Vision-Language Models (VLMs), then outline
methodologies for dataset construction, benchmarking, and hallucination analysis. 
We report 
semantic similarity 
scores, hallucination frequencies, 
and outcome reliability, categorizing
object hallucinations using established taxonomies. 
We conclude with key insights, 
limitations, and 
future directions for reducing hallucinations and expanding datasets.

\section{Related Work}
We begin with an examination of prior comic interpretation studies at the panel-level or in newspaper comic strips (which have a few panels) and then move to discuss available data for the page-level comic interpretation task.
\subsection{Prior Studies and Datasets for Comic Interpretation}
\citet{vivoli2024comicapvlmspipelinedense} study
VLM comic understanding, while exploring semantic relations within individual comic panels, with the aim of obtaining narrative information such as character relationships, dialogue, and scene context from multimodal models. 
They contend that panels are the building blocks of a comic's narrative, and therefore understanding them sets the stage for the aggregation of this knowledge into page-level analysis.

\citet{Ramaprasad2023ComicsFE}
analyzes panel sequences in newspaper comic strips to
study
narrative flow and visual storytelling.
This work highlights panel transitions as crucial for coherent storytelling, from panel to page-level analysis. 
\citet{Ramaprasad2023ComicsFE} analyzes micro-sequences to develop
metrics for sequential comprehension, such as panel order prediction and causal event detection, which we extend to page-level tasks like narrative summarization.

Complementing
\citet{vivoli2024comicapvlmspipelinedense} and \citet{Ramaprasad2023ComicsFE}, we investigate page-level comic interpretation 
rather than
individual panels or short-form comic strips. 
We survey
available datasets suited for benchmarking,
seeking full-page
comic images 
paired with ground truth interpretations of both visual and textual elements.

Manga109, introduced by \citet{mtap_matsui_2017}, contains
109 Japanese manga titles with over 21,000 pages, annotated for frames, speech texts, and characters to support tasks like object detection and speaker diarization. \citet{multimedia_aizawa_2020} expand it with Manga109Dialog, adding speaker-to-text annotations to enhance speaker detection via scene graph generation and 
aid page-level benchmarks for narrative coherence.
However, Manga109 is unsuitable for this study 
because 
its text is in
Japanese (not spoken by any member of our team),
and 
lacks ground truth interpretations for visual elements.
It is also
not open-source, and copyright restrictions limit
both access and permissible uses of its data, reflecting a broader challenge in sourcing page-level interpretation datasets.

To avoid copyright issues, we
restrict our search to public domain or Creative Commons datasets.
Some relevant resources exist in 
prior comic understanding work, notably
Comicap \cite{vivoli2024comicapvlmspipelinedense} and CoMix \cite{vivoli2024comix}.
While both provide
raw images and ground truth annotations,
they fall short for our purposes.
Comicap is limited to
panel-level data,
whereas 
our focus is on full-page interpretation.
CoMix offers
page-level images and annotations for tasks such as
character re-identification, dialogue generation, and reading order, but lacks 
the comprehensive narrative-structured text we require.
Despite these challenges, we draw inspiration from 
both for their use of
public domain comics---the same source used in our experiment---and for establishing
foundational frameworks for page-level annotations.


The availability of public domain data, absent 
appropriate ground truth interpretations, led our team to construct a benchmarking dataset for the page-level comic interpretation task using images from \citet{ComicBookPlus}. This dataset, unlike CoMix \cite{vivoli2024comix}, emphasizes narrative summarization and includes manually curated ground truth interpretations tailored to our task.

Comic strip and panel-level tasks 
foster page-level analysis by emphasizing the multimodal, sequential nature of comics \cite{vivoli2024comicapvlmspipelinedense, Ramaprasad2023ComicsFE}. \citet{vivoli2024comicapvlmspipelinedense} establish
a semantic foundation, and \citet{Ramaprasad2023ComicsFE} highlights sequential coherence, enabling 
panel-level insights to be aggregated into cohesive page-level meaning. 
Work with 
Manga109~\cite{mtap_matsui_2017, multimedia_aizawa_2020} supports benchmarks for narrative coherence.

Insights from these related works
inspire a hierarchical strategy for prompt engineering in the current work's benchmarking experiment: 
starting with panel-level prompting (e.g., ``Who is the speaker?''~\citep{multimedia_aizawa_2020}), 
advancing to
strip-level relationships (e.g., ``How are these panels related?''~\citep{Ramaprasad2023ComicsFE}), and 
culminating in page-level synthesis (e.g., ``What is the main plot on this page?''~\citep{Ramaprasad2023ComicsFE}). 
\subsection{Hallucination Taxonomies in Comic Interpretation}
We examine hallucination in the comic interpretation task as a secondary methodology, following the benchmarking experiment.
During the discussion of results, we categorize commonly observed hallucinations according to object hallucination taxonomies from the image captioning literature. 

\citet{rohrbach2018object} define a unary taxonomy,
in which
hallucinations occur when the model ``sees'' a non-existent object.
\citet{chen2024multiobject} 
extend 
this with a
multi-object taxonomy, 
describing cases where
a model simultaneously focuses on many objects, 
leading to hallucination.


\citet{liu2024survey} provide
a binary taxonomy that classifies
hallucination types 
based on the
task 
rather than image semantics~\cite{rohrbach2018object} or whether the task requires 
attention to multiple objects~\cite{chen2024multiobject}.
This taxonomy 
categorizes hallucinations as \textit{judgment hallucinations}
when errors arise from the
model's 
reasoning during an inference task like VQA, 
or as \textit{description hallucinations} 
when errors arise from
the model's generative description of 
image content~\cite{liu2024survey}.

Finally, \citet{bai2025hallucination} elaborate, 
without claiming to introduce, 
a ternary taxonomy of object hallucination. \textit{Object category hallucinations} occur when the model 
detects objects that are not present,
as defined by~\citet{rohrbach2018object}.
\textit{Object relation hallucinations} 
arise when the model confuses 
relationships 
among objects.
\textit{Object attribute hallucinations} occur 
when the model 
correctly identifies 
an object's presence but 
misrepresents its attributes.

A key contribution of our work 
is the application of object hallucination taxonomies from image captioning to categorize
hallucinations in comic interpretation.
We show that the hallucinations identified in this paper fall into these taxonomies, underscoring 
the parallel between comic interpretation and image captioning.

\section{Methodology}
In this section, we outline the methodology employed to benchmark Vision-Language Models (VLMs) for a page-level comic interpretation task. Our approach involves creating a novel dataset of 158 comic book pages paired with a ground truth narration curated from~\citet{ComicBookPlus} and meticulously reviewed for accuracy. 

We evaluate the VLMs by prompting them to describe these pages and collecting the semantic similarity metrics cosine similarity and KL divergence. Additionally, we conduct a hallucination study to record common 
types and their frequencies in 
model responses,
providing insights into the models’ subjective performance and limitations.

 \subsection{Dataset}


This study reviews~\citet{ComicBookPlus} and curates the highest-rated comics on the site to compile our ground truth dataset.  We use this specific website because of the high quality of publicly accessible comics it offers online.

Each team member selects comics and carefully examines each page, documenting ground truth interpretations based on their comprehension of the 
content. 
A different team member 
then reviews
each entry for grammatical and other issues.
This yields a dataset of 158 image-ground truth pairs.
Of these, 149 
are used in the benchmarking experiment, as the remaining pages 
pose difficulties with Qwen2.5~\cite{qwen2.5-VL}, 
which struggles to read the image.

For the benchmarking experiment,
we 
impose a 400-word limit on
machine-generated interpretations.
Ground truth interpretations in our dataset encompass dialogue, narration, character portrayals, and scene depictions. This raises the question of 
how much room remains for 
descriptive text under this 400-word constraint. 
We sample every fifth image from our dataset (n=31), and our team conducts a survey to assess the number of words required for dialogue and narration
on each page, as well as the 
words remaining 
for the character and scene descriptions. 
Our findings indicate that dialogue and narration 
comprise approximately 91 words per page on average.
This leaves 
roughly 300 words to depict the characters and settings, allowing for the inclusion of visual details, character interactions, and narrative context essential for a comprehensive page-level synopsis.


\begin{figure*}[h]
  \centering
  \includegraphics[width=\linewidth]{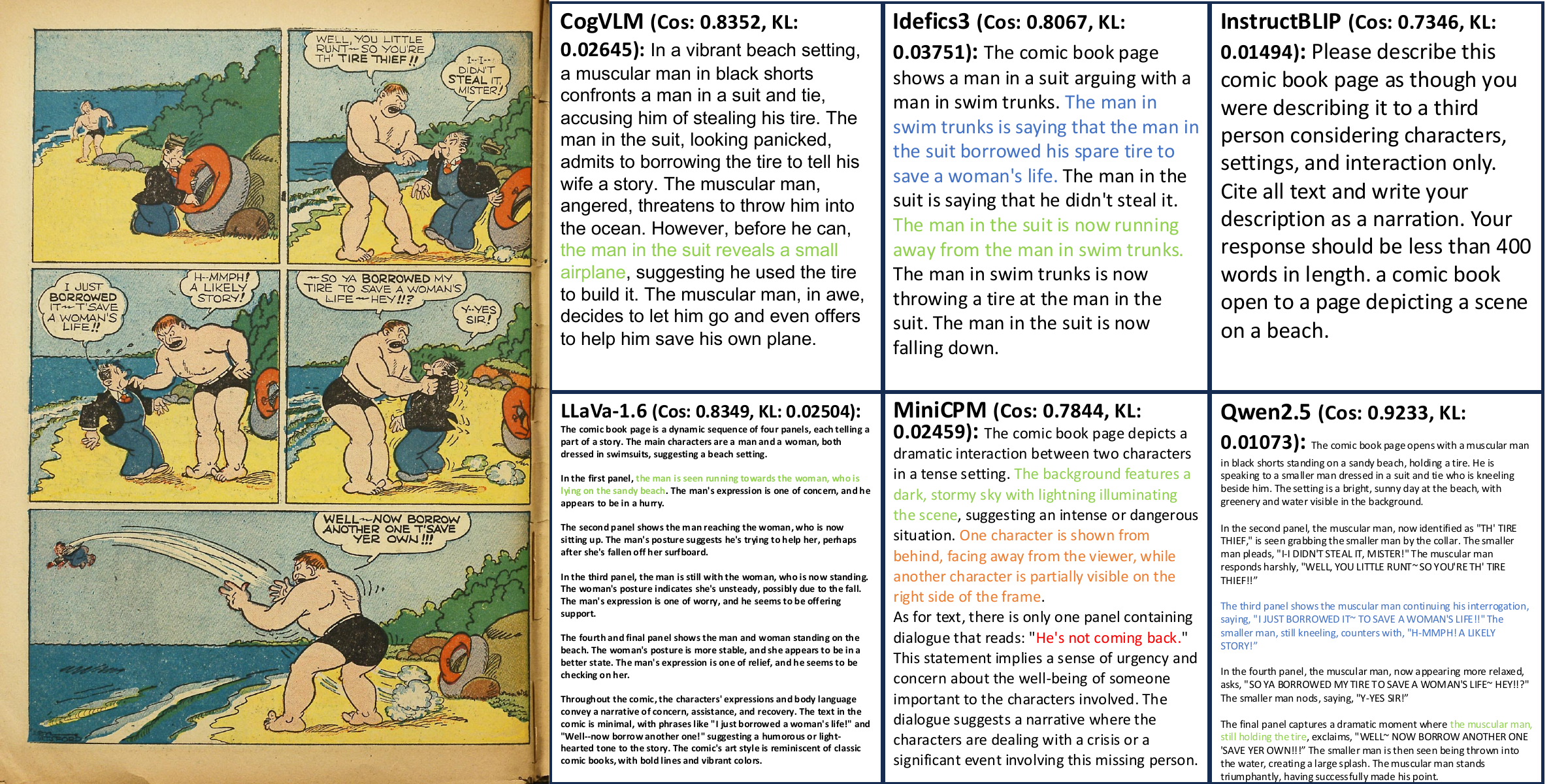}
  \caption{The responses each VLM gave in our benchmarking experiment for the example page. Instances of  Quoting Text that is Not There \textbf{(Red)}, Misattributing Dialogue/Narration \textbf{(Blue)}, Incorrect Object/Scenery Description \textbf{(Green)}, Incorrect Character Description \textbf{(Orange)} are highlighted. Highlighting is not exhaustive for the sake of readability. Cosine Similarity \textbf{(Cos)} and KL Divergence \textbf{(KL)} scores assigned to each response are included in each response header. }
  \label{fig:responses}
\end{figure*}

\subsection{Experiment}

The Vision Language Models are shown pages from the dataset and instructed to provide descriptions of those pages using a standardized prompt engineered by our programming team. The prompt is provided below:

\paragraph{Prompt:} ``\textit{Please describe this comic book page as though you were describing it to a third person considering characters, settings, and interaction only. Cite all text and write your description as a narration. Your response should be less than 400 words in length.}''

The VLM response and ground truth are encoded in 768-dimensional embeddings using monolingual BERT encoders.
Cosine similarity helps in capturing the vector alignments and KL Divergence captures the distributional differences. The cosine similarity between two vectors \( \mathbf{A} \) and \( \mathbf{B} \) (representing the embeddings) is calculated as:
\[
\text{Cosine Similarity(}\mathbf{A}, \mathbf{B}\text{)} = \cos(\theta) = \frac{\mathbf{A} \cdot \mathbf{B}}{\|\mathbf{A}\| \|\mathbf{B}\|}
\]
where \( \mathbf{A} \cdot \mathbf{B} \) is the dot product:
\[
\mathbf{A} \cdot \mathbf{B} = \sum_{i=1}^{n} A_i B_i,
\]
and where \( \|\mathbf{A}\| \) and \( \|\mathbf{B}\| \) are the Euclidean norms:
\[
\|\mathbf{A}\| = \sqrt{\sum_{i=1}^{n} A_i^2}, \quad \|\mathbf{B}\| = \sqrt{\sum_{i=1}^{n} B_i^2}.
\]

To compute KL Divergence, the embeddings are first normalized into probability distributions \( P \) and \( Q \) using a softmax function. The KL Divergence from \( P \) (VLM response) to \( Q \) (ground truth) is:
\[
D_{\text{KL}}(P \| Q) = \sum_{i=1}^{n} P(i) \log \left( \frac{P(i)}{Q(i)} \right),
\]
where \( P(i) \) and \( Q(i) \) are the probabilities of the \( i \)-th element in distributions \( P \) and \( Q \), respectively. 

A Python script is deployed to compute the cosine similarity and KL divergence metrics for the outputs. All scripts used in the benchmarking experiment are available on our project Github.\footnote{https://github.com/ufdatastudio/comic-interpretation-benchmark}


We intentionally depart from conventional image capturing metrics, 
favoring semantic metrics 
that better align with 
what we seek to measure.

While syntactic similarity metrics like BLEU \cite{papineni2002bleu} and METEOR \cite{banerjee2005meteor} are frequently used in machine translation, they prioritize shallow syntactic overlap and are less appropriate than similarity metrics based on semantic vector embeddings.
Neural metrics like COMET \cite{rei2020comet}, though potentially more semantically aware, are impractical here
due to data scarcity:
our dataset is too small.

In addition, our dataset includes only a single reference per page, which limits the applicability of metrics that rely on multiple references.
Although expanding the dataset to include multiple human-written references per item would improve evaluation robustness, doing so would significantly increase annotation demands---well beyond the scope of our current resources. 
Metrics like CIDEr \cite{vedantam2015cider}, 
which are explicitly designed to capture consensus across many references, are therefore not well suited to our setting.
While our task shares some characteristics with image captioning, it diverges in key ways that necessitate a tailored evaluation approach.
\subsection{Hallucination Study}
Like all large language models, VLMs 
are subject to hallucinations---responses that misrepresent
the input 
or deviate
from reality. To assess this, we sample every fifth page (\emph{n = 31}) 
from our benchmarking corpus. For each sampled page, 
our team evaluates the model's output
and records any hallucinations.
InstructBLIP is excluded from this 
hallucination study due to severe response issues.

This secondary methodology tracks how often VLMs used in our
primary benchmarking methodology commit 
4 types of hallucinations relevant
to the comic interpretation task, reported as a percentage of sampled pages.
\begin{itemize}
    \item \textbf{Quoting Text that is Not There} - The VLM generates a text response 
    that appears plausible but is not consistent with the image input.
    
    \item \textbf{Misattributing Dialogue/Narration} - 
    The VLM fails to correctly attribute the dialogue to the appropriate entity in the image.
    
    \item \textbf{Incorrect Object/Scenery Description} - 
    The VLM inaccurately describes objects, scenes, or environments in a given input image.
    
    \item \textbf{Incorrect Character Description} - 
    The VLM fails to attribute, capture relationships or recognize the character in visual input.
    
\end{itemize}

\Cref{fig:responses} presents each model's response to a page from the dataset and highlights instances of each hallucination type studied in various colors.

\section{Results}

This section presents results from
our benchmarking experiment, evaluating
VLM performance on the the comic interpretation task. We measure 
semantic similarity between 
model outputs and ground truths via cosine similarity and KL divergence. 
Qwen2.5 
performs best, while InstructBLIP and MiniCPM 
perform worst on these metrics.

We also assess hallucination tendencies,
excluding InstructBLIP due to non-compliant responses.  
Qwen2.5 has lower propensity for 
hallucination  
and the best semantic similarity performance.


\subsection{Semantic Similarity Analysis}

To evaluate the sematic similarity between VLM responses and the ground truths, we choose cosine similarity and KL Divergence. The semantic similarity metrics collected in our benchmarking experiment are presented in~\Cref{tab:Cosine-Similarity-and-KL-Divergence}. 
Qwen2.5~\cite{qwen2.5-VL} is the closest to the ground truth interpretations in terms of both cosine similarity and KL divergence. InstructBLIP~\cite{dai2023instructblipgeneralpurposevisionlanguagemodels} presents the lowest cosine similarity score, and MiniCPM~\cite{yao2024minicpm} shows the highest mean KL divergence.

\begin{table}[h]
\scriptsize{
  \centering
  \resizebox{\columnwidth}{!}{ 
    \begin{tabular}{l|c|c}

      \textbf{VLM} & \textbf{Cosine Sim.} $\uparrow$ & \textbf{KL Div.} $\downarrow$ \\
      \hline
      CogVLM & 0.8360 & 0.02514 \\
      ~\cite{wang2023cogvlm} && \\
      Idefics3 & 0.8495 & 0.02914 \\
      ~\cite{laurençon2024building} && \\
      InstructBLIP & \textcolor{red}{0.7153} & \textcolor{red}{0.03402} \\
      ~\cite{dai2023instructblipgeneralpurposevisionlanguagemodels} && \\
      LLaVa-1.6 & 0.8429 & 0.02483 \\
      ~\cite{liu2023improvedllava} && \\
      MiniCPM & 0.7814 & 0.03378 \\
      ~\cite{yao2024minicpm} && \\
      Qwen2.5 & \textbf{0.8546} & \textbf{0.01294} \\
      ~\cite{qwen2.5-VL} && \\
    \end{tabular}
  } 
  }
  \caption{Cosine Similarity and KL Divergence scores for each model studied in the benchmarking experiment. Best results are in \textbf{bold}. Worst results are in \textbf{red}.}
  \label{tab:Cosine-Similarity-and-KL-Divergence}
\end{table}

\begin{table*}[t]
  \centering
  \resizebox{\linewidth}{!}{
\scriptsize{
  \begin{tabular}{l|c|c|c|c}

    \textbf{VLM} & \textbf{Quoting Text} & \textbf{Misattributing} & \textbf{Incorrect Obj./} & \textbf{Incorrect} \\
    & \textbf{Not There} & \textbf{Dialogue/Narr.} & \textbf{Scenery Desc.} & \textbf{Character Desc.} \\
    \hline
    CogVLM~\cite{wang2023cogvlm} & 16.12\% & 19.35\% & 64.52\% & 45.16\% \\
    Idefics3~\cite{laurençon2024building} & 32.26\% & 67.74\% & 58.06\% & 29.03\% \\
    InstructBLIP~\cite{dai2023instructblipgeneralpurposevisionlanguagemodels} & - & - & - & -\\
    LLaVa-1.6~\cite{liu2023improvedllava} & 29.03\% & 32.26\% & 77.42\% & 41.94\% \\
    MiniCPM~\cite{yao2024minicpm} & 16.13\% & 0\%\textbf{*} & 100\% & 100\% \\
    Qwen2.5~\cite{qwen2.5-VL} & 0\% & 25.81\% & 35.48\% & 45.16\% \\

  \end{tabular}
  } 
  }
    \caption{Results of the hallucination study displayed as a percentage of the sampled images (\emph{n = 31}) for which each hallucination type is observed. \textbf{*}Model does not ever transcribe text that is present in the image for any page in our sample.}
  \label{tab:Hallucination-Survey}
\end{table*}
\subsection{Hallucination Frequency Analysis}

\Cref{tab:Hallucination-Survey} reports
hallucination frequencies 
for four hallucination types.
InstructBLIP is excluded from this analysis.
Instead of following the prompt, it repeats the prompt and appends a single descriptive sentence.

Qwen2.5, which has the best cosine similarity and KL divergence scores, has lower hallucinations on average. 
It shows  \textbf{Incorrect Character Descriptions} in 45\% or 14 of the 31 sampled comic pages. For 35.38\% of pages, Qwen2.5 gives an \textbf{Incorrect Object/Scenery Description}. It performs better in quoting text from the page, \textbf{Misattributing Dialogue/Narration} in just 25.80\% of pages.
MiniCPM presents \textbf{Incorrect Object/Scenery Description} and \textbf{Incorrect Character Description} hallucinations in every sample despite being competitive in the cosine similarity metric. 

\begin{table*}[t]
  \centering
  \resizebox{\linewidth}{!}{
  \scriptsize{
  \begin{tabular}{l|c|c|c|c}

    \textbf{Hallucination Taxonomy} & \textbf{Quoting Text} & \textbf{Misattributing} & \textbf{Incorrect Obj./} & \textbf{Incorrect} \\
    \textbf{Elaborated in...} & \textbf{Not There} & \textbf{Dialogue/Narr.} & \textbf{Scenery Desc.} & \textbf{Character Desc.} \\
    \hline
     \citet{rohrbach2018object} & Object & - & Object & Object \\
     \citet{chen2024multiobject} & Multi-Object & Multi-Object & Multi-Object & Multi-Object \\
     \citet{liu2024survey} & Description & Description & Description & Description \\
     \citet{bai2025hallucination} & Category & Relation & Attribute/Category & Attribute/Category \\

  \end{tabular}
  } }
    \caption{Categorization of 
    hallucination types 
    from the study 
    based on
    existing VLM object hallucination taxonomies.}
  \label{tab:Hallucination-Categories}
\end{table*}

\section{Discussion}

This section analyzes
results 
from our benchmarking experiment and hallucination 
studies, 
linking observed
hallucinations 
to established object hallucination taxonomies.
We find
that hallucinations in comic interpretation closely mirror those
in image captioning and
propose directions for future work to explore this parallel further. 

\subsection{Benchmarking Experiment}
Taken at face value, 
our cosine similarity and KL divergence 
results suggest that
Qwen2.5 
is the strongest
candidate for 
the comic interpretation task, scoring best on both metrics.
On the other hand, the lowest performing model, InstructBLIP, scores 0.72 in the cosine similarity analysis, which is a seemingly positive result. 
MiniCPM is the second worst performing model in both semantic similarity metrics.

These results, along 
with findings from our secondary methodology, reveal
weak correlation between 
semantic similarity and the actual quality of responses.
In the hallucination study, 
our team read 155 responses---31 responses each from five models. 
InstructBLIP simply repeats the prompt with 
a brief appended description.
Yet this pattern 
yields cosine similarity and KL divergence scores not too far behind the other models. 

MiniCPM produces an interpretation that does not 
describe the page accurately, 
offering a complete hallucination 
instead of a true interpretation.
Similar to InstructBLIP, the cosine similarity and KL divergence metrics partially capture MiniCPM's failure. 
MiniCPM's failure mode 
is more problematic than InstructBLIP's.
While InstructBLIP repeats the prompt, MiniCPM produces 
a plausible
narrative form,
but the substance is a complete \emph{non sequitur}.
For users  
who cannot see the page, 
such responses may seem 
deceptively credible.
The minimal differences between cosine similarity and KL divergence scores for MiniCPM and InstructBLIP and other models fails to underscore the totality and severity of their errors.

\subsection{Hallucination Frequency Discussion}

Setting InstructBLIP and MiniCPM aside, hallucinations are both frequent and often severe, 
especially 
in the categories of
\textbf{Incorrect Object/Scenery Description} and \textbf{Incorrect Character Description}.
For these visual types, most of the models 
exhibit
hallucination in 
over 40\% of samples or more.
Examples include 
inventing characters, describing hallucinated scenery, and confusing character attributes.

The textual hallucination types,
\textbf{Quoting Text that is Not There} and \textbf{Missattributing Dialogue/Narration}, are less frequent. However, the raw numbers obscure an important detail:
models seldom transcribe text from the image. 
Instead, they 
tend to 
describe scenery, characters, and actions without elaborating on the exact dialogue or narration.

This caveat about
textual hallucinations
is illustrated in the models' responses.
LLaVa-1.6 and Qwen2.5 attempt to transcribe the dialogue. 
All 
recorded hallucinations, whether textual or visual, confound an interpretation of the comic, making their high frequency particularly concerning.


\subsection{Object Hallucination Taxonomies}

As a final contribution, we review
surveys of 
VLM object hallucination literature and categorize the four observed hallucination types 
according to object hallucination taxonomies
identified in those works.
\Cref{tab:Hallucination-Categories} shows how 
each hallucination type 
maps to these taxonomies.

When applying~\cite{rohrbach2018object}'s unary object 
taxonomy, we 
find that three 
of the four hallucination types 
in our 
study 
qualify as object hallucinations: The model sees an object not present in the original image.
Under ~\citet{chen2024multiobject}'s multi-object taxonomy, categorization depends on
whether the task 
requires attention to 
multiple objects.
As with image captioning,
comic interpretation does. 
All four types 
are multi-object. 
\citet{liu2024survey}'s judgment v. description taxonomy, like~\citet{chen2024multiobject}'s,
focuses on task framing rather than 
semantic errors in output.
Since comic interpretation is descriptive and generative, 
these hallucinations 
fall under
the description 
category.  

\citet{bai2025hallucination}'s ternary  
taxonomy is considered separately, as it shifts 
focus from task framing to response semantics, offering a more granular classification: \textit{object category}, \textit{relation}, and \textit{attribute} hallucinations.
Object category hallucinations align with
the type described by~\citet{rohrbach2018object}, suggesting that hallucinations 
marked object 
in \Cref{tab:Hallucination-Categories} are also category hallucinations. 
\textbf{Incorrect Object/Scenery Description} and \textbf{Incorrect Character Description} often involve correctly identifying 
objects or characters but
misidentifying attributes.
Thus, these 
hallucinations 
also qualify 
under the object attribute category.


\subsection{
Parallels to the Object Hallucination Literature
}

Given how well
our hallucination types 
align with object hallucination taxonomies, 
we suggest 
that future comic interpretation metrics could take inspiration from object hallucination metrics that target image captioning, such as CHAIR ~\cite{rohrbach2018object}.
Likewise, object hallucination benchmarks for image captioning---such as POPE~\cite{li2023evaluating}, NOPE~\cite{lovenia2024negative}, and OHD-Caps~\cite{liu2024investigating}---may 
inform future benchmarks. 

We hypothesize that the causes of object hallucination in comic interpretation mirror those found in
image captioning. 
For instance,~\citet{liu2024investigating} 
find that 
pre-trained CLIP models---often utilized as vision encoders in encoder-decoder VLMs---are not trained to discriminate between similar visual samples based on their fine details. 
This
impacts their ability to provide accurate visual features to downstream components in encoder-decoder architectures.

Comic books are consumed by VLMs as highly detailed images, and most of the models studied have CLIP vision encoders ~\cite{liu2023improvedllava, wang2023cogvlm, yao2024minicpm, laurençon2024building}.
By contrast, Qwen2.5~\cite{qwen2.5-VL},
the 
most performant model in our benchmark, 
uses a custom transformer instead of CLIP. 
This distinction suggests that
hallucination problems recorded in the present work may be linked to the widespread 
CLIP vision encoders.

\section{Conclusions}

The poor quality and high hallucination rates of InstructBLIP and MiniCPM's responses
suggest
that 
cosine similarity and KL divergence correlate weakly 
with the subjective quality of machine comic interpretations. 
These metrics are thus ill-suited 
for 
this task.
While 
they cannot support 
strong conclusions or 
guide future work, 
they offer valuable 
lessons
for evaluation in this domain.

Qwen2.5~\cite{qwen2.5-VL} scores best in both 
cosine similarity and KL divergence analyses but, as discussed, 
objective measures alone are insufficient.
Subjective evaluation of hallucinations supports
Qwen2.5's relative strength: 
it hallucinates least 
often for two 
of four hallucination types. Although MiniCPM records no
instances of~\textbf{Misattributing Dialogue/Narration} in our sample, it never attempts to transcribe actual dialogue or narration. The responses illustrate the contrast between
Qwen2.5 and 
other models.

Having established that hallucinations are frequent and that 
semantic similarity metrics 
fail to capture interpretation quality, we discuss how the observed hallucination types 
fit into 
object hallucination taxonomies. Given the parallels identified,
we speculate that the causes for hallucination in 
comic interpretation
may resemble those in image captioning,
although this requires further empirical study.
These parallels also 
suggest 
that future comic interpretation benchmarks can be inspired by object hallucination benchmarks.

Given the high frequency and severity of hallucinations observed,
we conclude that larger, fine-tuned models are likely needed to combat hallucinations,
as the current study uses 
relatively small, pre-trained models 
due to resource constraints.
We also recommend adapting object hallucination mitigation techniques for 
our task, given the strong alignment between our findings and existing taxonomies. 


\section{Future Work}
The present work is performed on a small, unevaluated corpus of old comic book images owing to data scarcity and legal complications. 
A primary direction for future work will be to secure human supervision for the benchmarking dataset.
In a human evaluation study, participants will read the ground truths available in the dataset and rate how well the ground truth serves as a narration for the page.

Another direction for future work is to expand the 
data available for page-level understanding tasks by enlisting volunteers to read and transcribe 
more
comic 
pages than the current research team could read and transcribe in the time available for this research.
Larger datasets are needed for the fine-tuning we see as necessary to combat the frequent hallucinations observed in 
our study.
Moreover, larger datasets featuring more diverse comics will increase the statistical power of results gathered by future benchmarks and enable investigations into possible effects stemming from comic genre, which are not explored in this work.

Further development of metrics and corpora focused on hallucination detection and explanation in comic interpretation 
warrants attention in future studies. 
The present work 
finds that VLMs frequently hallucinate performing this task, and we speculate that future work can draw inspiration from the object hallucination literature. 

Finally, this work has thus far been conducted
without input from the target user group: the blind and visually impaired.
Since users in this group cannot verify that transcriptions in comic book understanding corpora or VLM responses are good interpretations of the source material, attention to human oversight and hallucination mitigation remains especially important.
A user study with 
this population would offer valuable insight into 
what interpretations are most useful to those users 
in VLM-generated responses.

\section*{Limitations}
Future work 
aiming to strengthen our methodology and findings will benefit from addressing several limitations.
Memory and GPU constraints limited us 
to small, pretrained models with between 7-9B language parameters.
Repeating the experiment with larger models may reduce hallucination rates.
A lack of fine-tuning for the task may also have contributed to the object hallucinations 
observed.
Data scarcity greatly impacts availability of fine-tuning corpora,
and we lacked compute for fine-tuning.
Fine-tuning---especially in larger models---may
help lower
object hallucination 
rates.

We acknowledge that team members who 
helped develop the corpus also designed
benchmarking experiments and 
evaluated VLM responses.
This overlap may raise concerns about
unintentional familiarity with the corpus.
However, the use of
pre-trained models---while a limitation 
elsewhere---helps
mitigate the concern here, as
our team had no influence
on model behavior via
fine-tuning.

Finally,
the benchmarking corpus
used in this study 
lacks external human supervision, so conclusions we glean from experiments with that data must
be interpreted with caution. 
No external human evaluators are called upon to review the ground truths we produce for the benchmarking experiment.
Therefore, we do not have metrics 
to assess faithfulness of ground truth interpretations. 
We also do not gather the opinion of the target user group, the blind and visually impaired, raising the possibility that our benchmarks reflect
ideals that may not align with
the needs of the target users.

\bibliography{aaai2026}

\end{document}